\documentclass[letterpaper, 10pt, conference]{ieeeconf}
\usepackage[T1]{fontenc}
\IEEEoverridecommandlockouts
\usepackage{cite}
\usepackage{amsmath,amssymb,amsfonts}
\usepackage{graphicx}
\usepackage{booktabs}
\usepackage{multirow}
\usepackage{array}
\usepackage{textcomp}
\usepackage{xcolor}
\usepackage[caption=false,font=footnotesize]{subfig}
\usepackage{url}
\usepackage[hidelinks]{hyperref}
\graphicspath{{figures/}}
\newcommand{\projecturl}{https://xrt-teleop.github.io}

\title{\LARGE \bf 
Evaluating Lightweight Contact Feedback in Dual-Operator Collaborative Humanoid Teleoperation
}

\author{Austin Graves$^{1}$, Chuizheng Kong$^{2}$, Yunho Cho$^{2}$, Yuanshao Yang$^{2}$, Advaith Nidumukkala$^{2}$,\\  Shreyas Kousik$^{2}$ and Mohsen Moghaddam$^{1}$%
  \thanks{This work has been submitted to the IEEE for possible publication. Copyright may be transferred without notice, after which this version may no longer be accessible.}%
  \thanks{$^{1}$Symbiotic and Augmented Intelligence Lab, Georgia Institute of Technology, Atlanta, GA, USA.}%
  \thanks{$^{2}$Robot Danger Lab, Georgia Institute of Technology, Atlanta, GA, USA.}%
  \thanks{Corresponding author: Austin Graves (\texttt{agraves@gatech.edu}).}%
  \thanks{Project website: \url{\projecturl}}}

\hypersetup{pdftitle={Evaluating Lightweight Contact Feedback in Dual-Operator Collaborative Humanoid Teleoperation},
  pdfauthor={Austin Graves, Chuizheng Kong, Yunho Cho, Yuanshao Yang, Advaith Nidumukkala, Shreyas Kousik, Mohsen Moghaddam}}

\begin{document}
\maketitle
\thispagestyle{empty}
\pagestyle{empty}

\begin{abstract}
Dual-operator humanoid teleoperation enables collaborative manipulation through two separate robots, but operators lack direct awareness of robot--object contact that is normally available during physical interaction. 
To address this, we present eXtended Reality Teleoperation (XRT), an extended reality teleoperation framework that returns a robot's fingertip contact through lightweight vibrotactile and/or auditory feedback, and a synchronized dual-humanoid testbed built on it for two operators.
Sixteen dyads performed a collaborative crate lift-and-flip under vision only and the three feedback conditions in a counterbalanced within-subject design with verified feedback-channel delivery.
Three results follow.
First, operators rated our finger-localized vibrotactile display as more effective than our open-ear sonification: all 32 ranked it above the sonification, 18 ranked it most effective, and its cues were rated far easier to perceive and interpret than the auditory cues.
Second, completion time, failures, workload, discomfort, and partner awareness showed no detected condition differences, with wide completion-time intervals and smaller meaningful effects unresolved.  
Third, performance changed substantially with practice, especially on the harder flip-over task: completion time improved across blocks, and attempt-to-attempt variation within a dyad (SD $\times$1.51) was larger than the largest condition contrast ($\times$1.16). Exploratory analyses detected no difference in immediate grip or motion responses between feedback conditions.
For novice collaborative teleoperation, our fingertip vibrotactile cues were rated more effective and easier to interpret than our auditory cues, while their effect on task performance remains unresolved.
\end{abstract}

\section{Introduction}

When two people carry a table, neither announces the plan. They can feel through the shared object: the small forces and motions that pass through a shared load tell a partner when the other has taken hold, is about to lift, and where the load is heading~\cite{sebanz2006joint,reed2008physical,groten2013intentions,mojtahedi2017communication}. This haptic channel supports coordination during shared-object manipulation tasks~\cite{reed2008physical,takagi2017physically}, and when it is removed, coordination relies more heavily on vision and speech~\cite{basdogan2000touch}. Teleoperating two humanoids removes direct physical sensation of robot--object contact. Each operator sees the task through a stereo video feed, but does not feel when the robot makes or loses contact with the object. Robot--object contact must therefore be inferred from visual motion and communication between operators. 
Lightweight contact feedback offers a way to restore part of this missing information by telling the operator which robot finger is in contact and when.
Headset-based humanoid teleoperation systems have largely been single-operator and without returned contact feedback; the closest two-operator XR humanoid system we are aware of returns no contact feedback and reports no dyad evaluation~\cite{zhao2026duet}.

\begin{figure}[t]
 \centering
 \includegraphics[width=0.99\linewidth]{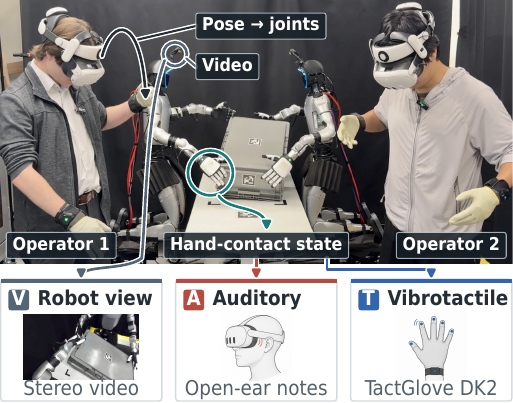}
 \caption{XRT signal paths for one station; the other mirrors them. Tracked pose commands the robot; camera video (V) and hand-contact cues (A/T) return to the operator. Robots kneel with fixed torsos for this tabletop task, without locomotion or balancing; only arms, hands, and neck are commanded. Photo staged with the study curtain drawn back; cue sketches are schematic.}
 \label{fig:teaser}
 \vspace*{-0.3cm}
\end{figure}

Contact feedback in headset-based humanoid teleoperation is usually absent~\cite{cheng2024opentelevision}; where provided, it is typically rendered as force through kinesthetic hardware, which improves force regulation~\cite{LiChappell2023,cuan2024leveraging,bergholz2023benefits} but closes a latency-sensitive bilateral loop and changes the operator's baseline~\cite{pacchierotti2024cutaneous}. We focus on a lightweight alternative: sensory substitution~\cite{MassiminoSheridan1992} of the robot's tactile skin through a commodity vibrotactile glove or a loudspeaker, adding no control loop and almost no hardware. The robots' skin reports contact and coarse activation, not calibrated force, so these cues convey \emph{contact state} (which finger, when, activation level, and when released). 
Such feedback may be easy to perceive and useful to operators without necessarily improving performance: prior teleoperation studies report cases where subjective preferences or workload changed without corresponding changes in task outcomes~\cite{gong2024airotouch,bremner2022impact}. This distinction is especially important here, where operators can also coordinate through vision and speech. We therefore ask not which modality works better, but how operators perceive and rate two lightweight displays of the same robot--object contact signal, and whether feedback that operators perceive and rate as effective changes measurable collaborative task performance.

We compare three feedback conditions, each added to the stereo video (V) that every operator has: our open-ear sonification (A+V), which plays audio cues through the headset's open-ear speakers and adds nothing to the operator's kit; our vibrotactile display (T+V), in which the robot finger that touched drives the tactile actuator on the operator's matching finger; and both (T+A+V). The combination is not a duplicate: the two cue types encode the same contact signal differently, the glove as a localized event on the finger that changed and the sonification as a note whose pitch names the finger and whose octave and loudness order the contact level, and they draw on separate perceptual resources~\cite{wickens2008multiple}; whether this redundancy helps or overloads is open~\cite{ohno2026beyond}. The three conditions plus the vision-only baseline form a $2\times2$ (tactile $\times$ audio) factorial.

\paragraph*{Proposed System}
To study lightweight contact feedback in physically coupled collaborative teleoperation, we built XRT, an extended reality teleoperation framework, and a dual-humanoid testbed on it (Fig.~\ref{fig:teaser}). In XRT, an operator drives a robot from a Quest~3 headset and TactGloves through a closed-form geometric retargeter~\cite{KongSEWMimic2026}, and one four-level quantizer of the robot's tactile-skin activation feeds both displays, so the cues share one contact signal and differ only in rendering; the framework is not tied to a particular robot. The testbed runs two XRT stations in lock-step on two Unitree~G1s with Inspire tactile-skin hands.

\paragraph*{Study Overview}
Sixteen dyads (32 operators) lifted and inverted a crate under all four conditions in a Williams square (128 attempts). Operators rated the vibrotactile display as more effective (all 32 ranked T+V above A+V, 18 ranked it most effective) and rated its cues far easier to interpret than our auditory cues. 
However, completion time, failures, workload, discomfort, and partner awareness showed no detected condition differences, and the approximate unadjusted minimum detectable completion-time effect was large (45\% at 80\% power), so smaller effects remain unresolved: the strong perceptual and perceived-effectiveness advantage of our vibrotactile cues did not translate into a resolved task-level advantage.  
Performance also changed substantially with practice: the harder flip-over sped up 19\% per block in the primary completion time and roughly halved over the session, and the attempt-to-attempt variation within a dyad exceeded the observed condition contrasts.

\paragraph*{Summary of Contributions}
(1) XRT, our extended reality teleoperation framework, with vibrotactile and auditory displays of one robot--object contact signal and time-aligned multimodal logging, and a synchronized dual-operator testbed built on it with two tactile-sensing humanoids and lock-step condition delivery.
(2) A 16-dyad within-subject study showing a strong distinction between perceived effectiveness and measured performance: novice operators consistently rated our vibrotactile cues as more effective and more easily interpreted than our auditory cues, while completion time, failures, workload, discomfort, and partner awareness showed no detected condition differences, with smaller effects unresolved.
(3) Evidence of a large within-session practice effect and execution variability in novice collaborative teleoperation, which bears on how contact-feedback studies must set operator training and attempt counts to resolve effects of plausible size.

\section{Related Work}\label{sec:related}

\subsection{XR Teleoperation of Humanoids and Avatars}
Humanoid teleoperation from commodity headsets is now well established. Open-TeleVision~\cite{cheng2024opentelevision} is a representative system with stereo video feedback; its authors note the absence of haptic feedback, and it was evaluated with four participants.
Our XRT builds on a closed-form shoulder--elbow--wrist (SEW) geometric retargeter~\cite{KongSEWMimic2026}; our testbed on it combines two elements we have not found together in headset-based humanoid teleoperation: two operators driving two humanoids on one object, and robot--object contact returned to each operator. The closest system we are aware of is Duet~\cite{zhao2026duet}, in which two operators control a G1 and a mobile manipulator through VR to collect demonstrations; it does not return contact feedback or report a dyad evaluation.

\subsection{Contact Feedback, Sensory Substitution, and Quality}
Vibrotactile or auditory displays of contact recover much of the benefit of force feedback~\cite{MassiminoSheridan1992}, and coarse low-frequency contact cues appear to carry much of that gain~\cite{wildenbeest2013impact}. On dexterous hands, binary vibration supported affordable teleoperation~\cite{coppola2022affordable}, and semantic wrist cues improved awareness and preference in bimanual tasks~\cite{huang2026semantic}. Auditory substitution includes per-finger sonification of a prosthetic grasp~\cite{Gonzalez2012}. The closest single-operator comparison to ours stacked vibrotactile and auditory transients on a force-feedback baseline and found them redundant yet complementary, with workload rising when combined~\cite{ohno2026beyond}. Force rendering itself changes force regulation: demonstrators with force feedback apply lower, less variable forces, and the benefit transfers to policies trained on their data~\cite{LiChappell2023,cuan2024leveraging,bergholz2023benefits}; that intervention is outside our scope. Subjective and objective measures can also dissociate: naturalistic tool vibration was preferred by 87\% of 30 operators without a workload change~\cite{gong2024airotouch}, and VR teleoperation sonification altered workload and stress but not task outcomes~\cite{bremner2022impact}. 

\subsection{Multi-Operator Teleoperation and Co-Manipulation}
Cooperative teleoperation of two humanoids transferring an object was proposed early with remote cockpits~\cite{Keerio2008}; that work concentrated on control design and did not report a user study or contact feedback. Human dyads coordinate better when they can feel each other through a shared object~\cite{reed2008physical}. To our knowledge, no prior study has tested in a controlled dyadic evaluation whether lightweight robot--object contact feedback that operators perceive and rate as effective also changes task-level outcomes in two-operator humanoid co-manipulation; this study provides that evaluation.

\section{System Description}\label{sec:system}

\begin{figure*}[!t]
\centering
\includegraphics[width=0.80\textwidth]{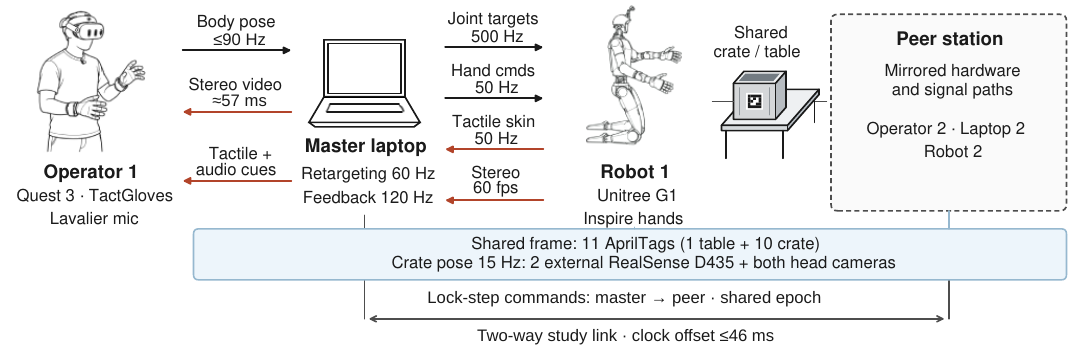}
\caption{The dual-humanoid testbed built on XRT, shown for the master station; the peer mirrors its hardware and signal paths. Both robots manipulate the shared tagged crate. Black arrows carry commands; red arrows carry sensing and feedback. The two-way study link carries master-to-peer lock-step commands, return tracking/status traffic, and clock synchronization (the 46\,ms shown is the maximum logged offset; median 2.1\,ms). Shared tracking is shown once; synchronized per-take logging is described in Sec.~III-E.}
\label{fig:arch}
\vspace*{-0.5cm}
\end{figure*}

XRT is an extended reality teleoperation framework: per station, a headset's body tracking is retargeted to a robot, and the robot's tactile-skin contact is returned to the operator through two feedback interfaces, a finger-localized vibrotactile glove and an open-ear sonification, that encode the same four-level contact state; neither the retargeter nor the displays are tied to a particular robot. For this study, two XRT stations form a synchronized dual-humanoid testbed (Fig.~\ref{fig:arch}) that acts on one object in lock-step.

\subsection{Hardware and Workspace}
Two Unitree G1 EDU+ humanoids kneel on height-adjusted scissor-lift platforms with their shins strapped down for stability; only the 7-DoF arms, hands, and neck are commanded. Liquid-cooled 50\,W Peltier coolers
limit shoulder-motor heating. Each robot carries Inspire RH56DFTP
tactile-skin hands and a two-axis Dynamixel neck gimbal with a ZED Mini
stereo camera. Operators wear a Meta Quest~3 with a facial fan to reduce
visually induced motion sickness~\cite{damour2017airflow}, and bHaptics
TactGloves DK2 with latex gloves underneath for hygiene and on top for improved tracking. 
Eleven AprilTags (a table world frame and a ten-tag crate bundle)
are tracked by two external RealSense D435 cameras and each G1's head
RealSense, fused at 15\,Hz. Laptops and headsets share a Wi-Fi~6E LAN for wireless operator control.

\subsection{Retargeting and Control}
The Meta Quest~3's onboard cameras provide markerless hand and
upper-body tracking.
The headset streams an 84-bone pose estimate over an unreliable WebRTC
data channel at up to 90\,Hz. Only the arms, hands and neck follow the
operator; the legs and waist remain fixed. At 60\,Hz, the laptop solves
each arm's seven joints from body-frame shoulder, elbow, and wrist
positions plus wrist orientation with the closed-form geometric solver SEW-Mimic of Kong et al.~\cite{KongSEWMimic2026}. Joint targets stream at 500\,Hz over DDS, finger angles from the tracked
hand at 50\,Hz over Modbus, and head yaw and pitch drive the neck. Safety
layers include a shoulder abduction clamp to avoid cooler interference, pose hold on lost tracking, current-triggered compliance, a head keep-out, an emergency stop, and a pause on headset removal.

\subsection{Video and Latency}
Each ZED Mini streams HD720 stereo at 60\,fps through a local relay into
a stereo shader (90$^\circ\times$60$^\circ$ field of view). Estimated video glass-to-glass latency averaged
57\,$\pm$\,3\,ms over 128 attempts (per-take p95 66\,ms) and teleop
motion-to-motion latency 87\,$\pm$\,9\,ms over 120 attempts (p95
142\,ms), from measured network, buffering, and actuation terms plus
assumed capture and display terms.

\begin{table}[!t]
\caption{Shared mapping of normalized tactile-skin activation (Sec.~III-D); downward band changes play the reversed bell an octave lower.}
\label{tab:mapping}
\centering
\footnotesize
\setlength{\tabcolsep}{3pt}
\begin{tabular}{@{}lcccc@{}}
\toprule
 & Light & Medium & High & Maximum \\
\midrule
Activation band & 0.10--0.35 & 0.35--0.60 & 0.60--0.85 & 0.85--1 \\
Glove amplitude (of 100) & 10 & 25 & 50 & 70 \\
Glove pulses (20\,ms) & 1 & 1 & 1 & 2 \\
Bell octave (upward) & 0 & $+1$ & $+1$ & $+2$ \\
Bell loudness (relative) & 0.10 & 0.35 & 0.60 & 0.85 \\
\bottomrule
\end{tabular}
\end{table}

\subsection{Contact Signal and Feedback Displays}
One robot-side signal, the normalized activation of the tactile skin,
drives both displays (Table~\ref{tab:mapping}). Each of twelve
regions per robot is reduced to a scalar (peak of fingertip and nail
patches; palm mean), baseline-subtracted, gated at a calibrated noise
floor ($\le$7 counts), and divided by the 1024-count ADC range. The
result encodes contact region and coarse activation, not force. Intensities stream to the headset
at 120\,Hz, where both displays quantize them into the same four bands.

The  TactGlove DK2 fires one burst on the affected finger per band change and
is otherwise silent.
The glove has fingertip and wrist actuators
but no palm actuator, so palm contact uses the wrist motor. 

Our sonification assigns each region a note on a perfect-fifth ladder
from palm to little finger:
\begin{equation}
f(n)=130.81\cdot2^{7n/12}\ \mathrm{Hz},\qquad n=0,\ldots,5.
\label{eq:sonification}
\end{equation}
A 125\,ms bell at the note of Eq.~(\ref{eq:sonification}) plays on each band change (exponential smoothing 0.80, hysteresis 0.05); notes sit on a fixed arc around the head with a per-finger timbre ramp. The notes came from the Quest~3's open-ear directional speakers with 3D spatial audio capabilities, so operators could communicate naturally and hear the robots; the notes were meant to be audible only to the wearer, although one operator reported occasionally hearing the partner's cues. Because the level at the ear varies with room noise, finger identity rides on pitch, and the contact level is coded in both octave and loudness; pitch and loudness changes interact perceptually~\cite{neuhoff2002pitch}, so a step in both is more salient but less separable.

\subsection{Lock-Step Control and Synchronized Logging}
A master laptop broadcasts study actions over TCP to its peer for
execution at a shared epoch 0.8\,s later. Attempts and conditions switch
together; logged clock offsets (median 2.1\,ms, maximum 46\,ms) align
the two records. Both headsets log their starting channel settings and
any later toggle. Per-robot records include joint state, goals,
frame-aligned tactile arrays ($\approx$50\,Hz), operator skeleton,
stereo video, feedback events, and latency. The master adds fused crate
pose (15\,Hz) and a manifest of condition and calibration. Speech was
transcribed live with a local Whisper ASR system and one lavalier microphone per operator; no audio recordings were kept. Offline, a MuJoCo twin replays both robots and the crate on the shared clock as a synchronized two-robot demonstration.

\section{User Study Design}\label{sec:study}

The study asks whether differences in how operators perceive and rate two lightweight displays of one robot--object contact signal translate into measurable task-level differences during physically coupled collaborative manipulation.

\subsection{Task and Conditions}
Pairs collaboratively lift a rigid plastic crate (61$\times$38\,cm footprint,
19\,cm internal depth, 1.68\,kg), invert it 180$^\circ$, and return it to the marked start area.
Each robot supports one side of the rigid crate: mismatched motions
pass through the object, and one partner must stabilize it against
sliding while the other rotates it. This makes the task physically coupled.
Each block held two recorded attempts: attempt~1 flips the upright crate
over (the \emph{flip-over}, rim grasp), attempt~2 flips it back (the
\emph{flip-back}, base grasp). The facilitator ended each attempt
once the crate was flipped and repositioned. Facilitator-called unfinished attempts count as failures and are assigned the 180\,s nominal limit as a failure penalty. Because these are not observed completion times, the primary model assigns them the limit rather than modeling censoring, and sensitivity analyses that use their observed stop times, exclude or inflate them, or apply censoring-aware rank tests left the conclusions unchanged (Sec.~\ref{sec:res-cost}).

Feedback has four levels: stereo video only (V), video plus our auditory
cues (A+V), video plus our vibration-based tactile cues (T+V), and all three
(T+A+V). The headset and both gloves were worn in every condition, for
consistent hand tracking, and the channels were toggled in software per
block, so the levels form a $2\times2$ tactile $\times$ audio factorial.
A+V needs no additional wearable and has precedent in sonified
contact~\cite{Gonzalez2012,bremner2022impact}; T+V is
fingertip-referenced; T+A+V tests whether combining these cues
protects detection under visual load or adds
workload~\cite{wickens2008multiple,ohno2026beyond} (Sec.~I). The two
displays encode the same contact bands with different spatial and temporal renderings (Sec.~III-D). They
substitute for contact sensation rather than render
force~\cite{MassiminoSheridan1992}: the robot's skin gives a
contact state, not a force magnitude, and coarse contact cues carry much
of the benefit of richer haptic feedback~\cite{wildenbeest2013impact}
without a bilateral loop~\cite{pacchierotti2024cutaneous}.

\subsection{Hypotheses and Pre-Specified Analyses}
The study addressed two hypotheses.
\textbf{H1:} in the collaborative dual-operator humanoid setting, each tested contact-feedback condition (A+V, T+V, T+A+V) yields higher task success and shorter completion time than the vision-only baseline.
\textbf{H2 (primary):} finger-localized vibrotactile display improves collaborative teleoperation beyond auditory display: A+V improves over V, and the vibrotactile conditions (T+V, T+A+V) outperform A+V on success and completion time. 
H2 also predicted that operators would judge our vibrotactile display more effective. That component was assessed by the final-survey item ranking the four conditions ``from most to least effective for completing the cooperative task''; there was no separate preference item. The protocol fixed
a linear mixed model with condition as a fixed effect and random
intercepts for dyad and for operator within dyad, and three ordered
contrasts, V vs.\ A+V, A+V vs.\ T+V (the central test of H2), and T+V
vs.\ T+A+V, with 95\% CIs and standardized effect sizes. Per-display perception ratings provided a complementary assessment of feedback quality.
Analysis procedures beyond the protocol model and contrasts, including
the dyad-level robustness checks, were developed post hoc.

\subsection{Participants, Design and Procedure}
We analyzed 32 operators in 16 dyads (22 aged 18--24; 27 novice robot
operators). Height was recorded in five bands spanning under 152\,cm to 190\,cm; the retargeter rescales operator geometry to the G1. Condition order followed a $4\times4$ Williams square~\cite{williams1949latin} with four dyads per sequence, balancing block position and first-order carryover; dyads were assigned to the four sequences in a fixed rotation by session order.  After a transcribed by-hand practice session of the crate flip to establish a communication and task approach baseline, operators stood leaning against a table to discourage torso movement (two later sat because of cybersickness, with reduced reach), behind a curtain that hid the robots and the crate, so the scene was seen only through the headset's stereo feed. Each
condition block comprised a per-finger demonstration of the block's display (each finger's cue triggered in turn, for the three feedback conditions), 2 minutes of free interaction with the feedback on, two recorded attempts (flip-over, then flip-back), then NASA-TLX~\cite{Hart1988}, a discomfort rating (0--10), and a partner-awareness item.  After all four blocks, operators completed the
System Usability Scale (SUS)~\cite{Brooke1996,bangor2009sus}, an
effectiveness ranking of the four conditions, eight perception items per
display (0--10), and satisfaction items.

\begin{figure}[!t]
\centering
\includegraphics[width=0.62\linewidth]{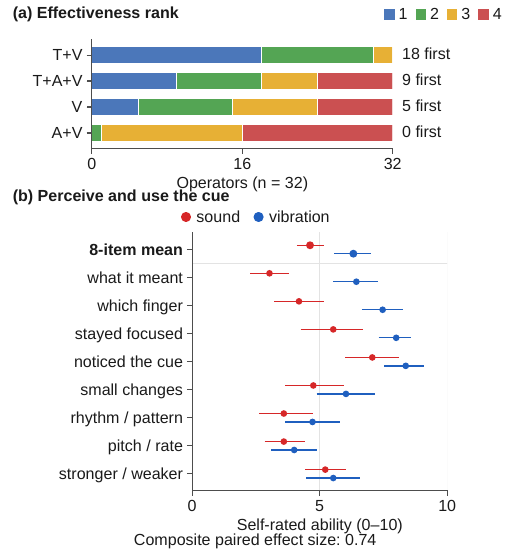}
\caption{Perceived effectiveness and interpretability (32 operators, 16 dyads). (a) Ranks from most (1) to least (4) effective for completing the task, with the number of operators ranking each condition most effective. (b) Perception-item means for our auditory (red) and vibrotactile (blue) cues; 95\% CIs resample complete dyads. The eight-item mean and its paired effect size $d_z$ summarize the display-specific advantage.}
\label{fig:pref}
\vspace*{-0.3cm}
\end{figure}

\subsection{Measures}
For completed attempts, completion time runs from crate-motion onset
(smoothed crate center $>$2.5\,cm from rest for $\ge$0.6\,s) to the
earlier of crate settle and last sustained contact plus 1\,s, on a
gated, Kalman-smoothed AprilTag track of the crate. Secondary time to flip
complete runs from the same onset to a crate tilt of at least
160$^\circ$ relative to its initial orientation, sustained for 0.5\,s. To explain the
outcome we also measured, from a MuJoCo digital twin,
which links touched the crate and when (any link within 3\,cm, released
at 5\,cm), and, around every tactile level-up, the operator's grip,
release, and hand motion in the following second.

\begin{figure*}[!t]
\centering
\includegraphics[width=0.62\textwidth]{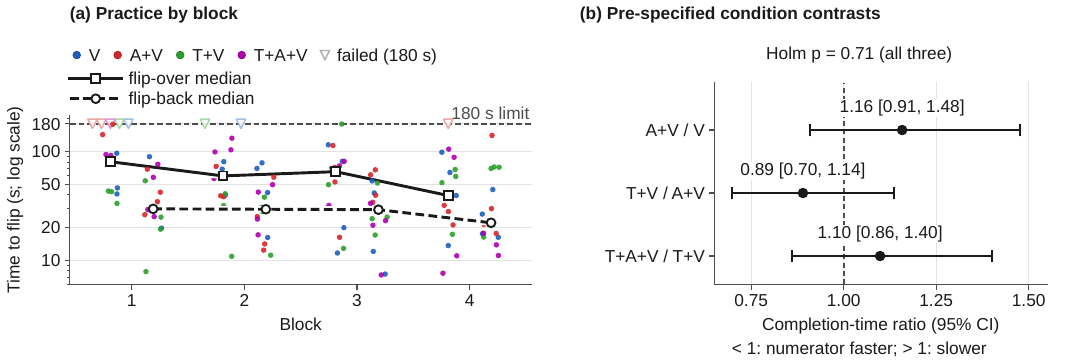}
\caption{No detected objective condition effect. (a) Time to flip complete (secondary outcome) by block and condition. Open triangles are the eight failures assigned 180\,s; squares on the solid line and circles on the dashed line are the per-block medians of finished flip-over and flip-back attempts. (b) Three pre-specified primary completion-time contrasts: mixed-model ratios and 95\% CIs, with the numerator condition listed first. The dashed line is equal time; Holm $p$ = .71 for all three.}
\label{fig:cost}
\vspace*{-0.4cm}
\end{figure*}

\subsection{Analysis}
Primary completion times are right-skewed and repeated within dyads. For attempt $f$ in block $b$ of dyad $d$, with time $T_{dbf}$ (s), we fit the linear mixed model
\begin{equation}
\begin{split}
\log T_{dbf} ={}& \beta_0 + \alpha_{c(db)} + \delta\,(b-\bar{b}) + \gamma f\\
                & + u_d + v_{db} + \epsilon_{dbf},
\end{split}
\label{eq:time-model}
\end{equation}
where $c(db)$ is the condition dyad $d$ received in block $b=1,\ldots,4$ ($\alpha_{\mathrm{V}}=0$, so $\alpha_c$ is the log time ratio of condition $c$ to V), $\delta$ is the practice slope per block ($\bar b = 2.5$), $\gamma$ is the flip-back ($f=1$) offset, and $u_d$, $v_{db}$ and $\epsilon_{dbf}$ are independent zero-mean Gaussian dyad, dyad--block, and residual terms; all coefficients are in log-seconds, so their exponentials are time ratios.
Objective operator outcomes add
operator-in-dyad intercepts. Subjective block
ratings use condition + linear block position with dyad and
operator-in-dyad intercepts. Covariates and dyad $\times$ block are
post-hoc additions. Restricted maximum likelihood (REML) supplies estimates and normal-reference Wald
95\% CIs; Holm adjusts exactly the three pre-specified
contrasts~\cite{holm1979sequential}. Omnibus condition likelihood-ratio tests (LRTs) use
maximum-likelihood full/reduced fits ($\chi^2_3$), never REML~\cite{pinheiro2000}.
Contrasts are time ratios $R_{c_1/c_0}=\exp(\widehat\alpha_{c_1}-\widehat\alpha_{c_0})$ from Eq.~(\ref{eq:time-model});
$R<1$ means shorter time in the numerator condition.

All 128 attempts, including eight failures, were retained. Finished
attempts contributed their measured times; failures were assigned
180\,s in both time models. Recorded measures end at the reviewed task endpoint; the offline endpoint review was not blind to condition. Event responses require
their full post-cue window before that endpoint; speech summaries use
complete utterances within the task window. Sensitivities use reviewed
observed durations, omit failures, assign 1.25$\times$ or 1.5$\times$180\,s, or
separate flip types. 

Exploratory log-time two one-sided tests (TOST)~\cite{lakens2017tost} on dyad means ($n$ = 16, $t_{15}$, 90\% CI) assess ratios in $[1/1.15,1.15]$. The approximate 45\% minimum detectable effect (MDE) uses 80\%
power and unadjusted two-sided $\alpha$=.05; multiplicity raises it.
 Failures use Cochran's $Q$.

Post-hoc subjective CIs bootstrap complete dyads (20,000 samples); Friedman uses 100,000 joint within-dyad label permutations, and the two T+V rank contrasts exhaust the $2^{16}$ within-dyad swaps (Holm over the two). These are exchangeability sensitivities; composite effect-size and grip-response CIs also bootstrap dyads, and the stratified log-rank/Gehan sensitivities permute whole blocks within dyads. Resampled and permuted time analyses use the assigned 180\,s for failures; alternative treatments are in Sec.~\ref{sec:res-cost}.

\section{Results}\label{sec:results}

\subsection{Rated More Effective and More Easily Interpreted}\label{sec:res-pref}
The effectiveness ranking was decisive (Fig.~\ref{fig:pref}a; Friedman statistic 37.7, $W$ = .39; dyad-permutation $p <$ .001). T+V was ranked most effective by 18 of 32 operators, T+A+V by 9, V by 5, and A+V by none. Every operator ranked T+V above A+V, consistent with H2's prediction; 23 of 32 ranked T+V above T+A+V (dyad rank-contrast Holm $p$ = .012). Our vibrotactile cues were rated far easier to perceive and use than the auditory cues (Fig.~\ref{fig:pref}b; eight-item means 6.32 vs 4.62, $d_z$ = 0.74, dyad-bootstrap 95\% CI [0.27, 1.48]), on every item, most of all for telling what a cue meant (6.44 vs 3.03) and which finger it referred to (7.47 vs 4.19). The free-text rationales agreed (dyad--operator codes): ``the haptic feedback helped isolate which fingers were making contact with the box'' (P12-1), whereas the sonification ``felt more like background music than anything else'' (P16-1). Self-reported force-awareness ratings were low: only 19\% agreed they could tell how much force they were applying, and 9\% how much their partner was.

\begin{table}[!t]
\caption{Outcomes by condition (16 dyads, 32 operators). LRT: mixed-model likelihood-ratio test.}
\label{tab:outcomes}
\centering
\footnotesize
\setlength{\tabcolsep}{3pt}
\begin{tabular}{@{}lcccc>{\raggedleft\arraybackslash}p{0.6in}@{}}
\toprule
 & V & A+V & T+V & T+A+V & Test, $p$ \\
\midrule
Primary time, median (s)$^{*}$ & 55.6 & 64.9 & 57.6 & 62.5 & LRT .55 \\
Failed attempts (of 32) & 2 & 3 & 2 & 1 & Cochran's $Q$ .78 \\
NASA-TLX (0--100) & 39.1 & 38.3 & 37.6 & 38.4 & LRT .95 \\
Discomfort (0--10) & 4.98 & 4.62 & 4.86 & 5.12 & LRT .60 \\
Partner awareness (1--5) & 3.75 & 3.81 & 4.00 & 3.78 & LRT .35 \\
Effectiveness rank (1 = most) & 2.62 & 3.47 & 1.50 & 2.41 & Friedman $<$.001 \\
Ranked most effective ($n$) & 5 & 0 & 18 & 9 & \\
\bottomrule
\multicolumn{6}{@{}l@{}}{\scriptsize $^{*}$Primary completion time; failures enter at 180\,s.}
\end{tabular}
\vspace*{-0.3cm}
\end{table}

\subsection{No Detected Objective Condition Effect}\label{sec:res-cost}
Delivery was verified: the feedback-event log matched the scheduled condition in 256 of 256 robot-takes, and cue rates were similar across conditions (70--77 level-ups per minute per robot; 22,182 events). Completion time did not differ detectably between conditions under the model of Eq.~(\ref{eq:time-model}) (Table~\ref{tab:outcomes}; LRT $\chi^2$(3) = 2.10, $p$ = .55; Wald $\chi^2$(3) = 1.97, $p$ = .58). The pre-specified contrasts, as time ratios of the richer over the poorer condition, were A+V/V $\times$1.16 [0.91, 1.48], T+V/A+V $\times$0.89 [0.70, 1.14], and T+A+V/T+V $\times$1.10 [0.86, 1.40] (Holm $p$ = .71 for all three; Fig.~\ref{fig:cost}b), so the completion-time components of H1 and H2 are not supported; their success components are non-evaluable at 120 of 128 successes. The protocol's simpler model (condition only, dyad intercept) gives the same point estimates with wider intervals (Holm $p$ = .91). Additional H1 contrasts against V were undetected (unadjusted 95\% CIs: T+V/V $\times$1.03 [0.81, 1.31]; T+A+V/V $\times$1.13 [0.89, 1.44]). No condition effect was detected under any of the failure-handling sensitivity analyses: entering them at their observed stop time instead, excluding them, inflating them, and fitting each flip type separately left every contrast within $\times$0.84--1.24 (Holm $p \geq$ .35), and the censoring-aware stratified log-rank and Gehan tests gave omnibus permutation $p \geq$ .91. Neither additive factorial main effect was detected (tactile $\times$1.00, $p$ = .97; audio $\times$1.13, $p$ = .17). Exploratory paired log-time TOST did not establish ratios within $[1/1.15,1.15]$ ($p$ = .36--.51): smaller, still meaningful effects remain unresolved, and we make no equivalence claim.

Workload did not differ detectably (TLX composite LRT $p$ = .95), nor did discomfort ($p$ = .60) or partner awareness ($p$ = .35); these non-significant tests do not establish equal workload or awareness. The exploratory height model found no detectable effect (dyad time $\times$0.82 per 10\,cm of the taller partner's height-band midpoint, $p$ = .08; height $\times$ condition $p$ = .96). Effectiveness and performance ranks showed little concordance: the per-operator Kendall $\tau$ between effectiveness rank and completion-time rank averaged +0.03, and the condition ranked most effective was faster than the lowest-ranked condition for 19 of 32 operators.

\subsection{Within-Session Practice Effect}\label{sec:res-learn}
A large within-session practice effect was observed (Fig.~\ref{fig:cost}a). Time to flip complete fell 25\% per block on the flip-over ($p <$ .001; finished-attempt medians 81, 60, 65, and 39\,s; model-based block 4 at 0.43 of block 1) while no flip-back slope was detected ($-$6\%, $p$ = .34; 30, 29, 29, and 22\,s); on the primary completion time the slopes were $-$19\% and $-$4\%. The practice effect did not depend detectably on the feedback condition (condition $\times$ block interaction $p$ = .34 for time to flip and .44 for the primary time; within each flip type, smallest $p$ = .080), and all eight failed attempts were flip-overs, five in the first block. The feedback point estimates were smaller: the largest, $\times$1.16, is less than one block of practice ($\times$0.75--0.81), less than the block-to-block scatter of a dyad's times under one condition (SD $\times$1.22), and far less than the attempt-to-attempt scatter within a block (SD $\times$1.51). This is a design observation, not an explanation of the null: a plateau was not established within four blocks, and the interaction tests have limited precision.

\subsection{Cue Coverage and Immediate Response (Exploratory)}\label{sec:res-mech}
Cue coverage, the fraction of digital-twin contact time with a tactile level registered, averaged 0.70 (condition range 0.69--0.73; LRT $p$ = .63); 29\% of contact time involved non-hand links. Non-hand contact may contribute to incomplete coverage, but these fractions do not establish its cause. For each of the 22,182 cue onsets we measured the whole-hand activation change over the following second (mean taxel sum in the second after minus the half-second before); 249 cues that fired within a second of the task end were excluded. Among the remaining 21,933, the change averaged +1,125 counts when vibrotactile output was enabled and +1,022 when disabled; auditory output could be present in either group. The within-operator difference was +103 counts (95\% dyad-bootstrap CI [$-$100, +295]), with no resolved immediate response difference. Matched control-time changes were near zero. Release and palm-speed comparisons were also inconclusive; none of the 16 exploratory operator-level tests survived correction, and overlapping event windows limit temporal specificity. Talk rate showed no detected condition effect (123 attempts with transcripts; LRT $p$ = .31). Higher talk rate was associated with longer completion time in a mixed model adjusting for block position and flip type, with dyad and dyad--block intercepts (elasticity +0.21, $q$ = .02 across 12 exploratory tests); causal direction is undetermined.

System usability was middling (SUS mean 60.4 $\pm$ 14.4 SD, ``OK''~\cite{bangor2009sus}) and correlated negatively with peak discomfort ($\rho$ = $-$0.36). Eighteen of 32 operators reported substantial cybersickness (discomfort $\geq$ 7 of 10), nine severely, with higher median workload (TLX 44.8 vs 29.8, $q$ = .02) and lower median usability (SUS 57.5 vs 68.8). Dyad time ($p$ = .77) and the condition ranked most effective ($p$ = .38) did not differ detectably between sickness groups.

\section{Discussion}\label{sec:discussion}

\textbf{A strong subjective difference without a resolved objective one.}
All 32 operators ranked our vibrotactile display above our auditory display, and average localization and interpretability ratings favored it. However, completion time, failures, workload, discomfort, and awareness showed no detected condition difference; smaller meaningful effects remain unresolved. Accordingly, H1 and the performance component of H2 were not supported. Subjective ratings and task benefit should thus be evaluated separately: prior studies likewise found preference~\cite{gong2024airotouch} or workload~\cite{bremner2022impact} shifting without outcome changes. 
The unresolved objective effect admits several explanations: any feedback effect may be smaller than this study could observe; vision and unrestricted speech may already support efficient coordination; the fingertip displays cover only part of the robot's contact; or the true effect on performance may simply be small. Our experiment cannot distinguish among these possibilities.

\textbf{Our open-ear sonification was hard to interpret.}
Operators rated its cues less interpretable, and none ranked A+V first. Pitch, timbre, and azimuth encoded finger identity while open-ear speakers competed with room sound. Decoding this mapping under load may require training~\cite{bremner2022impact}; the study did not isolate that mechanism. Combining the two cue types did not detectably improve partner awareness. This is an interface-level comparison between our finger-localized vibrotactile mapping and our open-ear sonification, not a modality-level comparison of haptic and auditory feedback. The results therefore motivate testing the mapping and combination under controlled listening conditions~\cite{ohno2026beyond}. 

\textbf{Within-session practice as a design consideration.}
Primary flip-over completion time fell about a fifth per block and roughly halved over the session, with no detected condition interaction. This does not explain the null feedback result: the data remain consistent with smaller feedback effects and with none. With 27 of 32 operators new to robot teleoperation, performance changed throughout the comparison, while cybersickness limited familiarization; operators noticed this themselves: ``the haptics primarily just made me feel more directly in control, but the task seemed to only get easier just by my increasing experience and familiarity'' (P11-1). Future studies should therefore train toward a plateau, model practice explicitly, and collect more than two attempts per condition.

\textbf{Cybersickness and the operator experience.}
Cybersickness's associations with workload and usability do not identify its cause. Head-following stereo, latency, and torso tracking motivate tests of viewpoint decoupling and spherical re-rendering~\cite{stauffert2020latency,chen2022decoupled,schwarz2021spherical}.

\section{Limitations}\label{sec:limitations}

Sixteen dyads and two attempts per condition leave broad time intervals; multiplicity further limits sensitivity. The MDE is a power approximation, not an exclusion bound, and the tested $[1/1.15,1.15]$ ratios were not established.  Analyses beyond the protocol model and three contrasts are post hoc.

Flip-over and flip-back differ in grasp and always occurred in that order; practice claims concern the across-block series, which did not establish a plateau. Success was near ceiling (120/128), limiting the success hypotheses' resolution.

Several design factors also bear on the null. Our auditory display was a single sonification design (pitch-coded finger identity, open-ear loudness) that operators had time to familiarize themselves with but were not fully trained on, so its results concern this design, not auditory contact feedback in general. Partners spoke freely, and talk rate did not differ by condition, so speech may have carried the coordination that contact cues could otherwise have improved; a speech-restricted condition is needed to isolate their contribution. Cybersickness affected more than half of the operators to varying degrees, adding uncontrolled between-dyad heterogeneity. The effectiveness ranking is unblinded self-report (operators could tell which display they had), so demand characteristics cannot be excluded.

Camera-based tracking was vulnerable to occlusion. Hands degraded during the study, with thumb flexion/extension range decreasing with time, and operator 2's robot losing a thumb's abduction range outright due to an abrupt collision 56\% of the way through the study; hardware/exclusion sensitivities did not resolve a condition effect. Hands also closed slowly relative to the arms. The tactile skin covers only finger patches and the palm (six display regions per hand), so back-of-hand, wrist, and forearm contact, 29\% of contact time, produced no signal that any mapping could display; twin-based coverage also depends on 3/5\,cm proximity thresholds and a 2--3\,cm inverted-state tracking bias, limiting generalization beyond this implementation. Event windows overlap and operator-level event tests are approximate.

Discomfort was a single item~\cite{keshavarz2011fms}; retrospective block-level NASA-TLX may reflect discomfort and cannot resolve momentary load~\cite{hart2006tlx}.

\section{Conclusion}\label{sec:conclusion}

XRT is an extended reality teleoperation framework that returns a robot's tactile-skin contact state through two lightweight displays of one signal, a finger-localized vibrotactile glove and an open-ear sonification; on it we built a synchronized two-operator, two-humanoid testbed.
In a sixteen-dyad study, every operator ranked our vibrotactile feedback above our open-ear sonification, and average interpretability ratings favored it, but completion time, failures, workload, discomfort, and partner awareness showed no detected condition difference, and smaller, still meaningful effects remain unresolved.
Performance changed substantially with practice within the session, highlighting the need for feedback studies with novice teleoperators to account explicitly for training state.
The Unity VR application, teleoperation stack, MuJoCo twin, and synchronized-logging tools are open source and linked from the project website, \url{\projecturl} (study recordings are not public); future work will test trained operators with enough attempts per condition to resolve smaller effects, and alternative auditory mappings with greater familiarization.

\section*{Acknowledgment}
The study was approved by the Georgia Institute of Technology Central Institutional Review Board (protocol IRB2025-897), and written informed consent was obtained from every participant.
Initial figures and drafts of the Results were prepared with assistance from LLMs (Claude, Anthropic; Codex, OpenAI) and were reviewed, verified, and edited by the authors.

\bibliographystyle{IEEEtran}
\bibliography{references}
\end{document}